\documentclass[runningheads]{llncs}
\usepackage[T1]{fontenc}
\usepackage{graphicx}
\usepackage{array}
\usepackage{listings}
\usepackage{booktabs}
\usepackage{tabularx} 

\begin{document}
\title{T2MM: An LLM Supported Architecture For Inquiry-Based Modeling}
\titlerunning{T2MM: LLM Supported Modeling}
%
\author{John Kos\inst{1} \and
Rudra Singh\inst{1} \and
Ashok Goel \inst{1}}
%




%
\institute{Georgia Institute of Technology, Atlanta GA 30322, USA}
%
\maketitle              
\begin{abstract}
Model Construction is a foundational practice in science learning that relies on visualization and interactivity.
Large Language Models, increasingly augmented with multimodal capabilities, have been integrated in education contexts to support learning.
However, these tools lack visual interactivity that is required by some learning contexts.
We introduce Text to Multimodal Model (T2MM), a robust, dynamic LLM supported architecture that assists in model construction within the open inquiry ecology-based modeling software Virtual Experimental Research Assistant (VERA).
T2MM accounts for the current context of the learner's model and creates interactive models, rather than static images, enabling the model to remain responsive to manual adjustment.
To measure technical feasibility, we evaluate T2MM through a custom procedurally generated dataset of natural language learner modeling requests and target models within the VERA system.
T2MM outperforms a baseline model generation architecture implemented through LLM-supported full code generation, common in the literature, across all measured success metrics.
Our contribution not only outlines LLM integration into a inquiry-based learning modeling tool, but also describes a possible architecture through which more interactive multimodal LLM tools can be created.

\end{abstract}
\section{Introduction}
Recent technological developments and the promise of adaptable personalized learning have led to the introduction of Large Language Models (LLMs) into a variety of learning environments across many subjects such as Coding \cite{li_coggen_2025}, Writing \cite{mou_willm_2025}, Mathematics \cite{gupta_beyond_2025}, and Scientific Inquiry \cite{hou_llm-enhanced_2025}.
Due to the nature of LLM technology, LLM integration defaults to a solely textual form of interaction \cite{bewersdorff_taking_2025}.
In the AI and education space, newer literature has introduced multimodality into LLM assisted learning environments, as visual representations are a necessary component for many different types of learning.
Examples of this include using images for question answering \cite{taneja_towards_2025}, and automated mathematical plotting \cite{christie_agentic_2025,bulusu_mathviz-e_2024}.

One form of multimodality, Modeling, specifically model construction, critique, and refinement, is a foundational practice used in science education, and inquiry-based learning \cite{vanlehn_model_2013,meadows_thinking_2008}.
Further description of inquiry-based learning is given in Section 2.1.
This form of science learning, based on the cycle of inquiry, presents a unique challenge for LLM integration as it is iterative and interactive in addition to being visual.
Existing multimodal architectures for education either do not attempt to implement multistep interactivity or when they do, they often struggle with technical issues \cite{christie_agentic_2025,bulusu_mathviz-e_2024}.
These technical issues will be discussed more in Section 2.
This culminates as a default to purely textual interaction for LLM integration into science learning and inquiry-based learning within the field AI and education \cite{hou_llm-enhanced_2025}.

Our work outlines an LLM assisted architecture, Text To Multimodal Model (T2MM), that receives textual requests from learners, and scaffolds model creation within an ecology focused conceptual modeling and simulation platform, VERA.
To our knowledge, this is the first AI-in-Education architecture to use LLM generated action sequences grounded in a live learner model state to support interactive conceptual modeling.
We evaluate T2MM using a custom procedurally generated dataset of 975 textual model creation requests, and target model pairs.
As a baseline, we created two alternative architectures, a 0-Shot prompting method (0-Shot) and N-Shot prompting method (N-Shot).
These two other architectures for model creation depend on LLM generated code, common in other parts of the literature \cite{yu_genai-drawio-creator_2026,bulusu_mathviz-e_2024}.
We demonstrate that T2MM outperforms the baselines across all success metrics for evaluating the structure of the model including: successful compilation, graph edit distance between the generated model and the target model, success based on size of the model the learner requested (as measured by length target action sequence), and success based on type of learner modeling request.

\section{Background}
\subsection{Model Construction For Science Learning}
Model construction is a core exercise in science education, as it models the process of inquiry in miniature \cite{vanlehn_model_2013,wilensky_thinking_2006}.
As part of their practice, scientists are able to construct models and iterate through a scientific process, using conceptual modeling and simulations \cite{bridewell_interactive_2006}.
For this reason, science education has implemented Interactive Learning Environments \cite{biswas_learning_2005,bauer_analysis_2017} where students can construct scientific models.
This allows science in education to mirror science in practice \cite{bybee_achieving_1995,flick_meanings_1993}.
Science learning however, specifically project-based or inquiry-based learning methods, require interactivity for the learner.
Built from constructivist theory, this takes the form of hypothesis generation, testing, and then hypothesis evaluation, which are repeated and described as the cycle of inquiry \cite{an_cognitive_2022}.
Key to this research, is the concept that completing all of the steps of scientific inquiry is necessary for learning gains \cite{kuhn_education_2005}.
VERA, an example of one of these Interactive Learning Environments, is built on an ecology-based ontology useful for science learning, which allows for students to complete the full cycle of inquiry. 
It follows that a tool preventing one of these steps would impact learning gains.

\subsection{Multimodal LLM Architectures}
Recent architectures such as MuDoc \cite{taneja_towards_2025} implement multimodality into LLM assisted learning environments.
MuDoc retrieves images from textbook PDFs, and utilizes them as supplements for question answering.
In a review by Bewersdorff et al. which describes possible applications of multimodal LLMs for science learning, diagram (read model) creation is listed as an expected avenue for application \cite{bewersdorff_taking_2025}.
However, no literature within the field of AI and Education has investigated LLM created interactive models.

While prior systems \cite{christie_agentic_2025,bulusu_mathviz-e_2024} have demonstrated LLM-assisted automated plotting for mathematics learning, the technical limitations documented in this work restrict their applicability to science learning contexts, where interactive, iterative model construction is central.
Research by Christie et. al. \cite{christie_agentic_2025}, uses LLM assisted function calling to display Desmos plotted graphs based on textual mathematical requests from the learner.
This system relies on full plot generation, lacking context awareness of the current plot displayed to the learner.
Work from Bulusu et al. \cite{bulusu_mathviz-e_2024}, also investigates automated plot generation, but additionally relies on a pipeline including Wolfram Alpha to bolster Desmos plot generation.
While this system is context aware, considering the current state of the learner plot, multistep actions are left to future work.
Both systems compare their architectures to full code generation as a baseline, either explicitly as part of the methods or implicitly in the discussion.

Research on automated model creation exists outside the field of AI and education \cite{yu_genai-drawio-creator_2026}.
The authors develop a pipeline using LLMs to generate the XML for Draw.io, which uses the same MxGraph package as VERA.
Additionally, they develop a simple automated XML validation tool to correct malformed XML returned by the LLM.
Unlike the above architectures, they create fully interactive models for the learner.
However, the system (1) implements model construction through code generation which was shown to be unreliable \cite{christie_agentic_2025,bulusu_mathviz-e_2024}, and (2) is only evaluated on a sample size of 10 examples.
Further, the ecology-based ontology of VERA presents a specific challenge for code generation as it requires more than properly formatted XML.
Nevertheless, we use a code generation engine as a comparison baseline.


\section{Methodology}

\begin{figure}[h]\label{T2MArc}
  \centering
  \includegraphics[width=.9\linewidth]{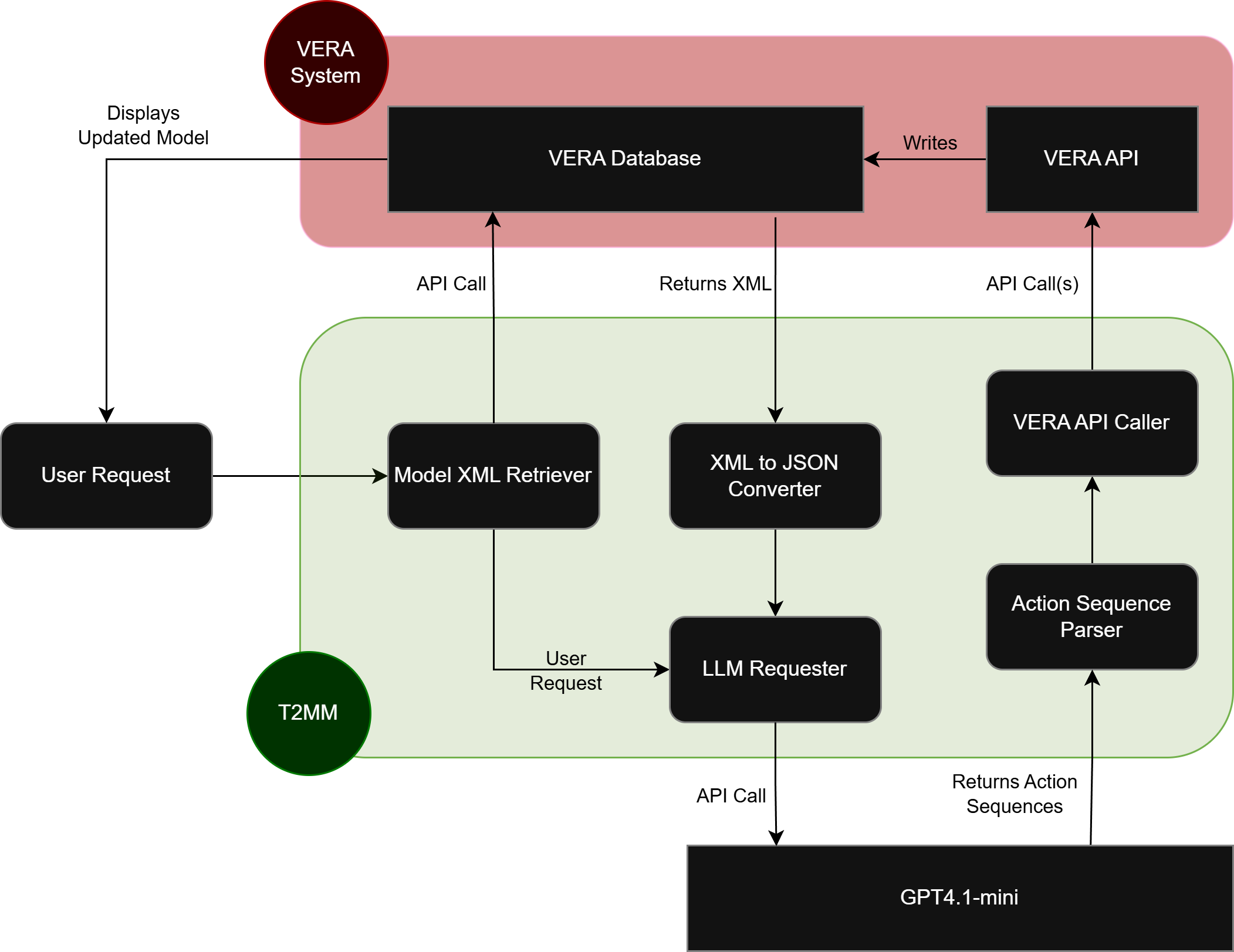}
  \caption{T2MM Architecture Diagram}
\end{figure}

Our architecture is outlined in Figure 1.
At a high level, the system consists of three major components:
\begin{enumerate}
    \item \textbf{VERA System}:
    The VERA system is outlined in Section 3.1. The T2MM system interacts with it in three ways: (1) the learner requests the creation of a model through the VERA chat interface, (2) the VERA database is queried following the learner request and the XML model representation of their model is returned to T2MM, and (3) T2MM then makes VERA API calls based on the LLM output. The API then writes to the model in the database, and the front end is updated for the learner.
    \item \textbf{T2MM}:
    T2MM is further discussed in Section 3.2. In short, it requests the XML from the VERA database, converts it to JSON and sends that JSON, along with a prompt and the learner textual request to the LLM. T2MM then receives \textit{action sequences} back from the LLM system which describe different actions that need to be taken on the current model to create the target model, e.g. \textit{Create Node}, \textit{Edit Relationship} or \textit{Delete Node}. These action sequences are checked against the ontology validator, to ensure that the actions are allowed under the VERA ontology, and if so, they are sent as requests to the VERA API which builds the model.
    \item \textbf{LLM System}:
    The LLM current system is ChatGPT 4.1-mini. It is accessible through the OpenAI API. GPT 4.1-mini was chosen as at time of analysis, it offered the most suitable price to expected performance ratio.
\end{enumerate}


\subsection{VERA}

The Ecology Modeling and Simulation System (VERA) [Citation Anonymized for Review] is an online ecology-focused conceptual modeling software.
Figure 2 represents a model in VERA.

\begin{figure}[h]\label{VERA}
  \centering
  \includegraphics[width=.9\linewidth]{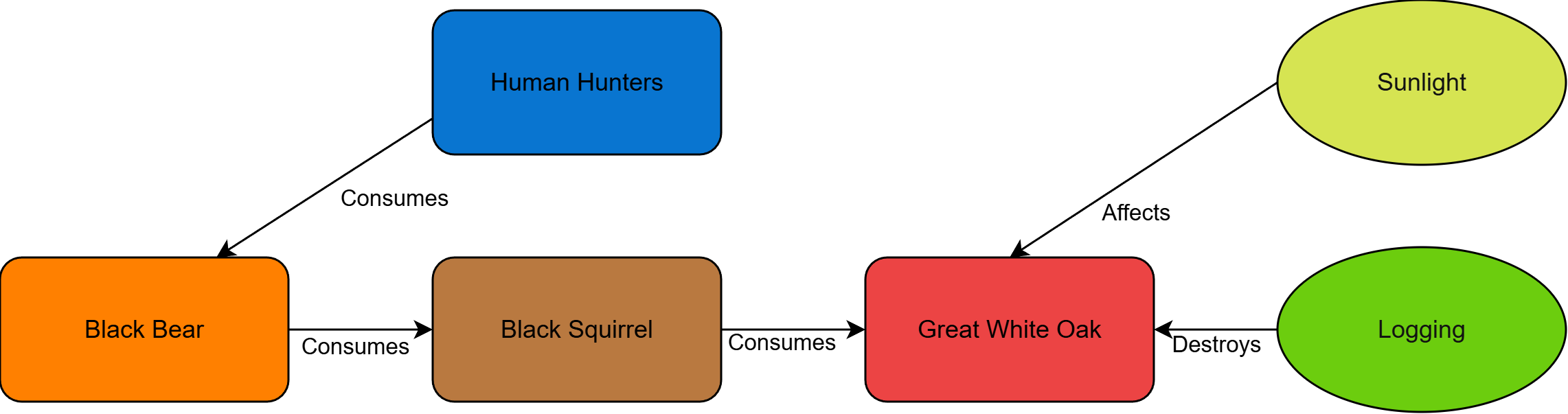}
  \caption{An VERA model representing the effects of hunting and logging on an ecosystem}
\end{figure}

Models are built using the Java MxGraph package, similar to Draw.io.
Models are then compiled [Citation Anonymized for Review] into a NetLogo \cite{tisue_netlogo_nodate} backend for simulation, which represents the different abiotic (e.g. rain, sunlight) and biotic (e.g. plants, animals, fungi) population levels in the form of a time series.
The front end presents the learner with a canvas with nodes representing biotics and abiotcs which they are able to drag and drop onto the canvas. 
Each biotic node has thirteen possible parameters representing descriptions such as: lifespan, carbon biomass, or number of offspring.
Between any combination of biotic and abiotic nodes, the learner can draw any of five relationships: Affects, Destroy, Becomes on Death, Consumes, and Produces.
Each of these relationship also has up to two of six parameters that change the interactions between the nodes including: interaction probability and consumption rate.
All of this data, which describes a model, is represented in the VERA database through XML.
Fully constructed models are too verbose to include in this paper, but for clarity of understanding, it is helpful to visualize them as a classical computer science graph where the nodes and edges both contain parameters.
Below is the XML representation of an empty model:
\begin{lstlisting}[frame=single, basicstyle=\ttfamily\small]
    <mxGraphModel><root><mxCell id="0"/><mxCell id="1"
    parent="0"/></root></mxGraphModel>
\end{lstlisting}

\subsection{T2MM Architecture with example}
The internals of the T2MM Architecture are outlined in Figure 1.
\subsubsection{Model XML Retriever}
The Model XML Retriever pulls the XML, like the example shown above, from the VERA Database.
\subsubsection{XML to JSON Converter}
The XML is then converted to a minimally sized JSON representation. In the case of an empty model, an empty set of nodes and relationships is sent. A simple model is given below.
\begin{lstlisting}[frame=single, basicstyle=\ttfamily\small]
"nodes": [
{"id": 1, "name":"lichen", "type":"biotic", "properties":{}},
{"id": 2, "name":"frog", "type":"biotic", "properties":{}},
],
"relationships": [
{"id":3, "source":"lichen", "target":"frog", "source_id":1,
"target_id":2, "type":"Destroys", "properties":{}},
]
\end{lstlisting}
\subsubsection{LLM Requester}
The converted JSON is combined with the learner request and a prompt outlining the VERA ontology.
The learner request, corresponding to the model given in this section, for example: \textit{``Remove the node lichen and its relationships''.}
The prompt also contains N-Shot examples of learner request, model JSON, and expected LLM actions returned.
\subsubsection{Action Sequence Parser}
The action sequence parser reads through the action sequences returned by the LLM, which takes the form of the code below.
Each of these \textit{actions} map to a change that can be made to a given model.
\begin{lstlisting}[frame=single, basicstyle=\ttfamily\small]
[{"action":"remove_relationship", "id":3},
{"action":"remove_node", "target":"lichen", "target_id": 1}]
\end{lstlisting}
The action sequence parser reads through each of the actions and validates that they are acceptable within the VERA ontology.
Specifically, it ensures that parameters for biotics are properly named, and that the relationships between biotics and abiotics is acceptable, e.g. it does not make sense to say an abiotic consumes a biotic.
These structured commands serve as a series of steps for constructing or editing the conceptual model and can be directly interpreted by the VERA backend.
\subsubsection{VERA API Caller}
The action sequences are then mapped to the VERA API which then writes the XML to the VERA Database.
Each action maps to a backend method such as \texttt{addNode()}, \texttt{addRelationship()}, or \texttt{updateParameter()}, which update the internal model representation. 
This ensures all model updates create properly formatted XML which can be compiled by the VERA system. The updated model is then rendered on the VERA canvas in real time.

\subsection{Dataset}
The dataset is made up of two parts, the text description of a request to the LLM, and JSON code generated from expert models created by research scientists in biology.
Starting with the set of five expert XML models created by the research scientists, we took a number of steps to increase the size of the dataset, all which correspond to the set of actions a learner can take in VERA: (1) Node Creation or Deletion, (2) Relationship Creation or Deletion, and (3) Parameter Adjustment for Nodes or Relationships.
First we converted the expert models into our JSON format, which can be reconverted back to XML using the VERA API, to make manipulation more simple.
Second, we took the structure of the expert models, and manually decomposed them such that they were in various stages of creation, ranging from empty model to full model.
Next we substituted the various names and types of the different nodes, changing them from biotic to abiotic, or generating new names from a random pool.
Lastly, we algorithmically changed various parameters for the nodes and relationships.

To generate the natural language request, we tied each of the actions described above to a natural language phase.
For example, if the learner wanted to create a model consisting of a bear consuming an owl, the generated natural language request would be ``Add bear and owl, and make bear consume the owl.''
This gave us a sample of 975 model and natural language pairs.

The intention of this dataset was to evaluate if T2MM demonstrated the technical feasibility to create properly structured models, as the visual ontology of VERA is comparatively simple.
Additionally, due to this, our evaluation only investigates the correctness of the model structure.

\subsection{Evaluation and Comparison to 0-Shot and N-Shot XML code generation}
As a baseline, we compare T2MM with LLM supported 0-Shot and N-Shot XML code generation architectures.
The generation of complete code through LLM is either the baseline or the final implementation in the existing literature \cite{yu_genai-drawio-creator_2026,bulusu_mathviz-e_2024}.

For the 0-Shot and N-Shot architectures, the XML for an VERA model was sent to ChatGPT4.1-mini, and the LLM XML response was set as the model code in the VERA database.
The ontology description in the prompts for the 0-Shot and N-Shot architecture matched T2MM.
Prompts and other additional materials are hosted on OSF \footnote{https://osf.io/cr7hv/overview?view\_only=bea596d85cae408aa8269ea8a1b01b86}.
T2MM differs from the 0-Shot and N-Shot architectures as it relies on LLM generated action sequences, rather than an XML response.
Additionally, in the N-Shot case, XML examples made up of current model XML, textual requests from the learner, and expected XML responses were given for node creation, edge creation, and parameter changes.

The dataset is evaluated through four metrics, (1) compilation, (2) graph edit distance between the generated model and the target model, (3) success by size of learner request as determined by length of target action sequences, and (4) success by type of learner request.
Compilation is determined by evaluating the XML using both an XML parser, and through automated error checking with Selenium when the generated XML was set as the model data in the VERA database.
Failure to compile in either mode was counted as an XML compilation failure.
Graph Edit Distance is determined by converting both the action sequences returned by T2MM and the XML returned by the 0-Shot and N-Shot architectures into JSON.
We then evaluate if the generated JSON matches the target model JSON by comparing the structure.
A model is considered to have the same structure if (1) all the nodes have the same set of names with the correct types (biotic or abiotic), (2) if the relationships match by type (consumes, destroys, etc.) and correspondence between the correct nodes, and (3) if the parameters for the nodes and relationships match.
If all of these aspects match, the graph edit distance is considered zero, and the models are considered matching.
Any change to the structure of a graph would increase the graph edit distance.
For example, changing the value or name of a parameter would increase the graph edit distance by one, and changing the name of a node would increase the graph edit distance by one plus the number of corresponding relationships that include that node.
All models with a graph edit distance greater than zero are considered incorrect for future evaluation metrics.
Success by size of learner request was evaluated by using the graph edit distance and the length of the target action sequence.
Success by request type was determined by breaking learner modeling requests into three different types: \textit{Create Model}, \textit{Edit Model}, and \textit{Change Parameter}.
Create model is when a learner starts with an empty model and requests the creation of nodes or relationships.
Edit model is when the learner requests node or relationship changes to an existing model, changing the structure.
Change Parameter is when the changes requested by the learner only involved a change to parameters.

\section{Results}
We evaluated our system using four different success metrics.
First, we evaluated whether the XML created by T2MM, or returned by the two whole code generation methods successfully compiled.
Second we determined the graph edit distance between the models created by the architectures, and the target models.
In this case, a Graph Edit Distance of -1 indicates the model failed to compile, a Graph Edit Distance of 0 indicates the generated model matched the target model, and a Graph Edit Distance \(>=1\) indicates the number of actions required to make the generated model into the target model.
Third, the architectures were evaluated by the number of actions required to change the learner's current model, at time of modeling request, into the expected target model.
Lastly, we evaluated success by type of action the learner is requesting, whether that is a full creation of a model (Model Creation), Editing their existing model (Edit Model), or changing a parameter (Parameter Change). 

\subsection{Compilation}

As shown in Table 1, overall, the XML generated by all systems compiled near perfectly.
T2MM, unlike the architectures however, compiled in all cases which is necessary to ensure that learners engage in the cycle of inquiry uninterrupted.

\begin{table}[h]
    \centering
    \caption{Success at XML compilation by architecture}
    \begin{tabular}{lcc}
        \toprule
        \textbf{Architecture} & \textbf{Compiles \#} & \textbf{Compiles \%} \\
        \midrule
         T2MM & 975 & 100 \\
         0-Shot & 942 & 96.62 \\
         N-Shot & 972 & 99.69 \\
        \bottomrule
    \end{tabular}

    \label{tab:text_desc_example}
\end{table}

\subsection{Graph Edit Distance Between target model and Generated Model}
\begin{table}[h]
\centering
\caption{Graph Edit Distance between target model and generated model}
\begin{tabular}{c|c c c c c c c c c c c}
\hline
 & -1 & 0 & 1 & 2 & 3 & 4 & 5 & 6 & 7 & 8 & 9+ \\
\hline
T2MM & 0 & 904 & 5 & 22 & 0 & 18 & 0 & 19 & 0 & 7 & 0 \\
0-Shot  & 33 & 531 & 69 & 84 & 94 & 0 & 73 & 0 & 69 & 22 & 0 \\
N-Shot   & 3 & 560 & 49 & 115 & 16 & 26 & 15 & 16 & 24 & 12 & 139\\
\hline
\end{tabular}
\label{tab:counts_single}
\end{table}

Table 2 shows the graph edit distance between the target model and the generated model.
The averages and standard across architectures were: T2MM ($\bar{x} =$  0.298, SD $=$ 1.21), 0-Shot ($\bar{x} =$ 1.638, SD $=$ 2.34), and N-Shot ($\bar{x} =$ 3.572, SD $=$ 7.22) 
A Graph Edit Distance of -1 indicates cases where the XML did not compile and were removed from the averages.
N-Shot, which demonstrated the worst performance, generated models with graph edit distances of up to forty two edits.
Investigation of the generated JSON revealed that the N-Shot architecture would change parameters for nodes and relationships seemingly at random.
All of the parameter values given in the N-Shot examples were the default given during node or relationship creation, so it is unlikely that the example caused this parameter variation. 
Uninvestigated, however, is why the same level of random parameter variation did not occur for the 0-Shot case, as the prompt was the same, excluding the examples, and results were generated in the same run.

\subsection{Success by Length of Target Action Sequence}
Figure 3 illustrates the probability of success by length of target action sequence.
T2MM demonstrates a strong ability to build models of up to five actions in length.
Success is around 50 percent for seven actions, and below 4 percent for eight actions.
0-Shot and N-Shot are less than eighty percent successful at making a single edit to a model, and performance drops off  severely around three actions.
All architectures achieved full success for action sequence length of four, however there were only four cases in this category.

\begin{figure}[h]\label{edit}
  \centering
  \includegraphics[width=\linewidth]{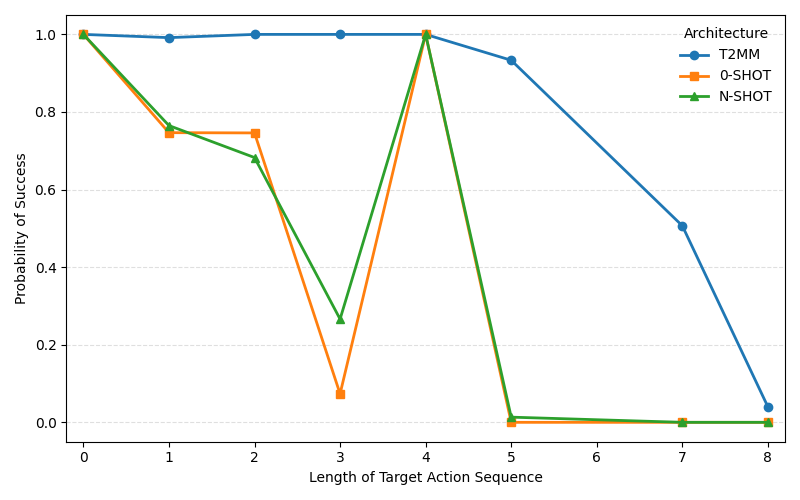}
  \caption{Probability of Success by Length of target action sequence}
\end{figure}

Table 3 shows the number of cases for target action sequence length. 
The majority of cases for the target action sequence was a single edit long.
Length of 0 refers to an instruction by the learner to not make changes to the model.

\begin{table}[h]
\centering
\caption{Total \# refers to length of target action sequence}
\begin{tabular}{c|cccccccccc}
\hline
 & 0 & 1 & 2 & 3 & 4 & 5 & 6 & 7 & 8 \\
\hline
Total \# & 24 & 600 & 63 & 109 & 4 & 75 & 0 & 75 & 25 \\
\hline
\end{tabular}
\label{tab:by_action}
\end{table}

\subsection{Success Across Models by Learner Request Type}

Figure 4 demonstrates the success rate by type of learner request.
For T2MM, create model had the lowest accuracy of 79.38\%.
Edit model and Change Parameters achieved much higher accuracy indicating that if T2MM fails to create the target model in the first iteration of a learner request, they have a high chance of success for subsequent attempts.
The 0-Shot and N-Shot architectures were largely unsuccessful at creating a model when starting from an empty canvas, with the 0-Shot architecture failing in all cases to create a model.
The N-Shot architecture was only slightly more successful successfully creating a model in 8.31\% of cases.
This gives some indication that the code generation architectures heavily rely on the current model XML to properly format changes according to the VERA ontology.
This is reinforced by the higher success rates for editing a model or changing a parameter.

\begin{figure}[h]\label{type}
  \centering
  \includegraphics[width=\linewidth]{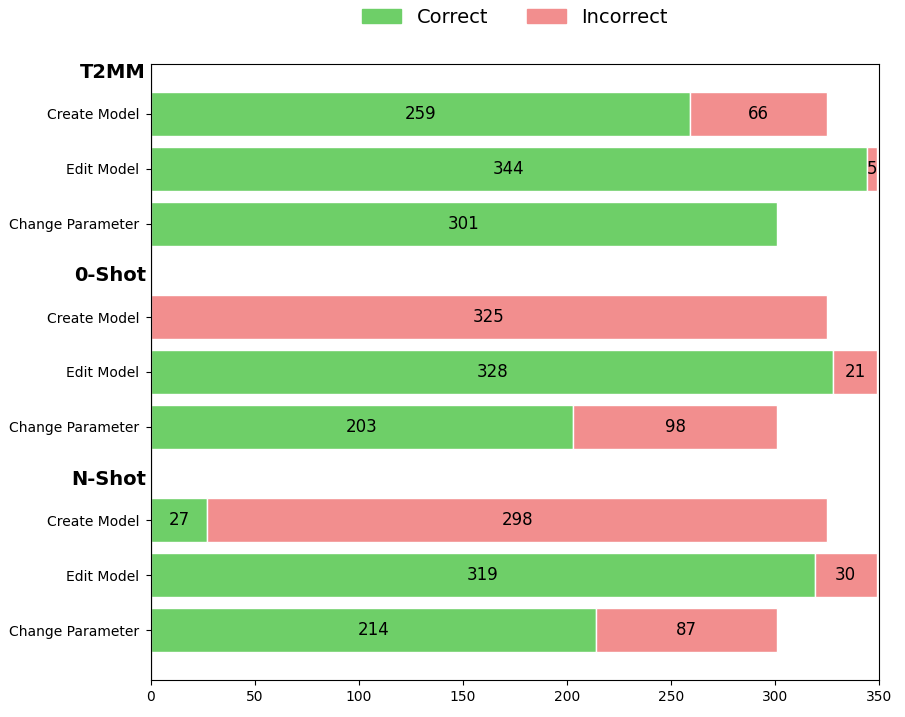}
  \caption{T2MM Success by action type}
\end{figure}

\section{Discussion, Future Work, Limitations}
Our results indicate that full XML generation for model creation is unsuccessful if the XML has to adhere to a specific ontology.
That is, The N-Shot and 0-Shot architectures exhibited stronger performance in cases where there was more XML in the VERA ontology to mimic, such as for model editing and parameter changes.
One possible explanation for this behavior is that model creation requires the generation of a large amount of XML text, which due to the architecture of generative AI introduces more opportunities for hallucination, or incorrect token generation \cite{zheng_why_2025}.
By comparison, our T2MM architecture depends on a much more compact JSON representation of the model.
This promises reliability, which is important for learners engaging in the cycle of inquiry \cite{an_cognitive_2022}.
This is further demonstrated by the failure of the 0-Shot and N-Shot architectures to create a model starting from an empty canvas.
If a learner were rely on either of these systems for assistance in building a model from scratch, they would struggle to engage in inquiry as it would improperly represent their hypothesis, and would be unsuccessful in iteration.

This also highlights the ability of T2MM in multi-step context aware model construction, which was a challenge outlined in the literature \cite{bulusu_mathviz-e_2024}.
As for why our architecture is better able to engage in iterative model construction, one reason is that our architecture is action focused rather than result focused.
Put simply, T2MM receives the \textit{actions} from the LLM that need to be taken on the current model to reach the target model, while existing literature expects the LLM to return fully formed representations of the expected outcome \cite{bulusu_mathviz-e_2024,yu_genai-drawio-creator_2026}.

Cognitive load theory provides another connection to this work.
Inquiry-based learning platforms, such as VERA, which is used to describe complex systems, are often challenging for learners to use due to the high level of cognitive load \cite{kirschner_why_2006}.
Previous research suggests that while pictures or conceptual models can help reduce cognitive load \cite{mayer_nine_2003}, especially when displaying complex systems \cite{bobek_creating_2016}, asking the learner to draw these complex systems raises the cognitive load \cite{leutner_cognitive_2009}.
LLM assisted model creation, could reduce the cognitive load in model creation, allowing the learner to struggle less with inquiry-based learning.

While this work presents an architecture for interactive LLM scaffolding within a science learning environment, there are a handful of limitations.
Firstly, this architecture is focused on the ontology of the VERA system and other science learning environments may not have similar technical requirements, whether that is XML based models, or a preexisting API that supports the ontology of model construction.
For this reason, generalization of the architecture is an open question, however we hope that evidence for the reliability of the system can be used by other researchers.
Additionally, this research only evaluated these architectures based on the graph structure of the model built in the VERA system.
Aspects such as the visual representation of the models were not evaluated.
However, compared to other systems in the literature, the visual representations in the VERA system are relatively simple, such that they can be implemented programmatically.

Left to future work is full evaluation by expert users.
This includes textual requests created from actual learners of the VERA system, and qualitative evaluation of the created model including: appropriateness of the plot to the learning context, and whether it matches the textual description.
Additionally, textual requests by real learners will present a unique challenge for the robustness of the system.
Lastly, additional future work includes expert evaluation of the visual aspects of the T2MM system.

\section{Conclusion}
In this work, we develop an interactive LLM architecture that supports multi-step learner modeling within a science learning environment.
Our main contribution is the outline of an architecture for model construction that is reliable, and can support ontologies specific to science learning.
This multimodality and interactivity, novel to the AI and education space, makes our architecture well crafted for supporting the cycle of inquiry.
We compare our research to LLM supported XML full code generation, using a custom procedurally generated dataset of 975 learner textual requests and model pairs.
We find that despite the success of full code generation in creating well structured XML output, our system is far more successful at interaction with existing knowledge representations, which is necessary for model editing.
This evaluation isolates representational correctness and action sequencing as prerequisites for learner-facing deployment, leaving linguistic variability and visual evaluation to future work.

\subsection{Acknowledgments}

This research has been supported by NSF Grants \#2112532 and \#2247790 to the National AI Institute for Adult Learning and Online Education.
We thank members of the VERA project in the Design Intelligence Laboratory for their contributions to this work.

%
%
%
\bibliographystyle{splncs04}
\bibliography{references}
\end{document}